\documentclass[journal]{IEEEtran}
\usepackage{amsmath,amsfonts}
\usepackage{algorithmic}
\usepackage{algorithm}
\usepackage{array}
\usepackage[caption=false,font=normalsize,labelfont=sf,textfont=sf]{subfig}
\usepackage{textcomp}
\usepackage{stfloats}
\usepackage{url}
\usepackage{verbatim}
\usepackage{multirow}
\usepackage{graphicx}
\usepackage{xcolor, soul}
\usepackage{color}
\sethlcolor{yellow}
\usepackage[center]{caption}

\begin{document}

\title{Accurate Cutting-point Estimation for Robotic Lychee Harvesting through Geometry-aware Learning}

\author{Gengming Zhang, Hao Cao, Kewei Hu, Yaoqiang Pan, Yuqin Deng, \\ Hongjun Wang$^{\#}$, and Hanwen Kang$^{\#}$ \\[2pt]
\small{College of Engineering, South China Agriculture University, Guangzhou, China}\\
\small{$^{\#}$ Correspond Authors}}

\maketitle

\begin{abstract}
Accurately identifying lychee-picking points in unstructured orchard environments and obtaining their coordinate locations is critical to the success of lychee-picking robots. However, traditional two-dimensional (2D) image-based object detection methods often struggle due to the complex geometric structures of branches, leaves and fruits, leading to incorrect determination of lychee picking points. In this study, we propose a Fcaf3d-lychee network model specifically designed for the accurate localisation of lychee picking points. Point cloud data of lychee picking points in natural environments are acquired using Microsoft's Azure Kinect DK time-of-flight (TOF) camera through multi-view stitching. We augment the Fully Convolutional Anchor-Free 3D Object Detection (Fcaf3d) model with a squeeze-and-excitation (SE) module, which exploits human visual attention mechanisms for improved feature extraction of lychee picking points. The trained network model is evaluated on a test set of lychee-picking locations and achieves an impressive $F_{1}$ score of 88.57\%, significantly outperforming existing models. Subsequent three-dimensional (3D) position detection of picking points in real lychee orchard environments yields high accuracy, even under varying degrees of occlusion. Localisation errors of lychee picking points are within ±1.5 cm in all directions, demonstrating the robustness and generality of the model.
\end{abstract}

\section{Introduction}
Precision agriculture or precision farming is a modern concept of whole-farm management that uses a variety of technologies ranging from remote sensing and proximal data collection to automation and robotics. In this study we are looking at the harvesting of fresh lychees. This work has proved to be very challenging, not only in terms of the mechanical design of the robot, but also in terms of the vision system, navigation, control methods and manipulation system \cite{r65}. In this case, our main focus is the detection of picking points on lychee stems in point cloud information to improve the autonomous harvesting of traditional picking robots.

Single fruits such as apples and oranges can be picked directly, while cluster fruits such as lychee and grapes need to be picked for the whole cluster. Lychee can only be picked by identifying the main fruit-bearing branch (MFBB) picking point and then cutting the MFBB to prevent fruit damage \cite{r2}. If the workspace is clear and fruits are not blocked by obstacles, it is not challenging to detach these fruits off trees \cite{r63}. However, the field is unstructured, and lychee fruit bunches can be of different sizes in shape and appear at different heights and positions, in addition, the lychee picking points in unstructured scenarios are prone to dense occlusion (or the target position is not obvious enough), and small targets are difficult to detect, thus affecting the detection accuracy. Thus, it has become the focus of research to find accurate and directed locating of lychee fruit bunch picking points, and two key technologies are needed to achieve this demand: (1) three-dimensional (3D) point cloud data that resist occlusion, and (2) a 3D point cloud detection CNN model with the ability of precise location under the premise of the former.

In recent years, computer vision-based technologies have been widely used to locate picking points for fruit \cite{r3,r42,r4}. The characteristics of the target object itself (such as color, shape, texture, etc.) are used for lychee detection and identification through image processing (e.g. image filtering, image segmentation, morphological processing) and machine learning algorithms and such operations \cite{r5,r43,r6,r7,r44}. With the profound development of deep learning, especially convolutional neural networks (CNN), which further enhances the generalisation ability as well as the robustness of machine vision perception systems, more and more researchers have turned to 2D image detection methods based on deep learning detection. Object detection algorithms, including You Only Look Once (YOLO) and deep learning networks improved on YOLO \cite{r11,r12,r45,r46,r47}, and a two-stage detection represented by the RCNN (region-based convolutional neural networks) series \cite{r50}. The above study achieved good results for the detection of picking points, but it is still difficult to challenge the situation due to light variations, occlusion by leaves and branches, etc.

The advantages of 3D point cloud data itself (e.g. accuracy in representing geometric shapes, easy access to depth information, 3D reconstruction helps to improve robustness and accuracy of detection) as well as the intensive development of 3D convolutional neural networks have motivated researchers to explore point cloud-based methods for fruit harvesting and picking point detection. Based on the idea that point cloud reconstruction is not limited by perspective and helps to improve the robustness and accuracy of detection, pure manipulation of point cloud data has also become a popular method for fruit picking point detection \cite{r51,r16,r52,r53,r14,r54,r66,r57}. Although the above research methods have achieved satisfactory results, there are still significant research gaps in lychee picking point detection.

In this work, a robotic arm-assisted data processing based on TOF camera point cloud fusion and Fcaf3d-lychee network model for lychee picking points was proposed, which can be used to accurately locate lychee picking points in the natural orchard environment, and was tested in the field. The proposed method fused multi-view point cloud data from a TOF camera, and then visually perceived the fused sensor data to find the exact location of the lychee picking point. The results show that the results of this study have higher localisation accuracy than the typical 3D point cloud target detection model. In summary, the following contributions are presented in this paper:

\begin{itemize}
    \item A model Fcaf3d-lychee is proposed for lychee picking point detection, which greatly improves the accuracy of lychee picking point localization.

    \item TOF camera for multi-view point cloud stitching, effectively solving the problem of single-view occlusion and too little sensory information.
    
    \item Demonstration of the lychee picking robot equipped with the Fcaf3d-lychee model on recognition and localization of picking points in the natural orchard environments.
\end{itemize}

The rest of this paper is organised as follows. Section \ref{section: review} surveys related work. The system overview and methodologies of our approach are presented in Section \ref{section: method}. The experiment results and discussion are presented in Section \ref{section: experiment}, followed by the conclusion in Section \ref{section: conclusion}.

\section{Related Works} \label{section: review}
\subsection{Review on Image-Based 2D Target Detection in lychees}
Currently, the application field of robotics is extensive, covering many basic and related technology areas \cite{r17}. Among them, in the intelligent development of harvesting robots, vision algorithms have been a key factor affecting the overall performance of the system. The development of vision algorithms for fruit harvesting robots has two main tasks: target localization as well as picking point extraction, and the challenges include target color distortion caused by variable natural light environments, interference of interlaced plant organs on recognition algorithms, and intraclass deformations of fruits of the same species \cite{r18}. The use of machine vision and its related algorithms has targeted solutions to improve efficiency, accuracy, intelligence, and remote interaction during robot operations \cite{r19}. To categorize them, the early major ones are single-feature analysis methods, multi-feature fusion analysis methods, and pattern recognition methods \cite{r20,r21}. \cite{r24} proposed a recognition method based on an improved linear discriminant analysis (LDA) classifier to address the problem of the low success rate of lychee-picking robots in recognizing green lychee (not fully ripe lychee) due to background interference. The core is an LDA method for extracting image convolutional features, and the idea of "maximum margin" in the SVM algorithm is introduced to determine the threshold value of LDA, which is finally integrated into a multiple LDA classification system by the Adaboost method. Through experiments, the accuracy of this classifier for unripe lychee is 80.4\%, and the algorithm can also be applied to the determination of fruit ripeness. The above research is based on image processing, which has high requirements for light and still needs to go through a certain calculation process when dealing with the 3D position of the picking point, resulting in poor picking efficiency and poor robustness.
\subsection{Review on Deep Learning-Based lychee 2D Target Detection Methods}
In the existing research, stereo vision measurement techniques and traditional digital image processing techniques have been adopted by many researchers, while the booming development of deep learning in the field of image processing provides a better solution for fruit recognition, thus enabling fruit-picking robots to adapt to complex orchard environments \cite{r25}. To improve the ability of the picking robot to operate throughout the day, an algorithm for the detection of lychee fruits and fruit stalks for nighttime environments was proposed \cite{r26}. The performance of the algorithm under different artificial light intensities was tested separately, and the YOLOv3 convolutional neural network and U-Net network were utilized to detect fruits and fruit stalks, and the model finally performed best under normal brightness, achieving an average detection accuracy of 99.57\% as well as segmentation effect of 84.33\% MIoU. However, the method first uses YOLOv3 to obtain the location of the stringed fruit and then utilizes the ROI region for fruit stalk recognition, which is not efficient to execute. \cite{r28} improved the network structure of YOLOv5s for the background occlusion and fruit occlusion scenarios that are easily encountered when picking apples so that it can autonomously determine the objects that can be grasped and those that cannot be grasped during the picking operation. The method, however, is mostly applicable to single fruits, or for estimating yields, and do not apply to lychee harvesting scenarios. The above studies have made great progress in the acquisition of fruit-picking points by target detection through 2D images, but most of the studies have not investigated the direct detection of mother branch picking points.
\subsection{Review on Deep Learning-Based 3D Target Detection Methods for Fruits}
Although target detection via 2D images has made great progress in picking points for fruit-picking robots, most of the studies have not investigated the direct detection of mother branch picking points. While 2D images may lack detailed spatial distribution information, 3D point clouds obtained by depth camera methods provide a more comprehensive representation of object distribution, especially in complex scenes with occlusions and overlaps. The potential of 3D point clouds has been increasingly recognized, resulting in new methods for object detection. \cite{r33} proposed a strategy for accurate picking point localization of artificially cultivated grapes by combining near and far and utilizing a combination of depth point cloud data. Firstly, the grape clusters were roughly detected for localization in the far-view image, and then in the near-view image, the data features of the depth point cloud and the structural features of the horizontal lattice grapes were fully utilized, and 95 out of 100 samples were successfully localized, with an accuracy of 95\%. The localization accuracy was as high as 95\%. In addition, \cite{r36} proposed a pomegranate tree organ classification and fruit counting method based on multi-feature fusion and support vector machine (SVM) to address the limitation of occlusion in the traditional 2D image-based recognition method, firstly, a 3D point cloud is obtained, then the point cloud is preprocessed, and its color and shape features are extracted for training, and a classification model is obtained, and experiments show that this method can detect most of the fruits on the tree. Existing point cloud target detection methods, however, have not been studied for picking points.

Our work is based on the Fcaf3d model, which is improved to get the 3D detection model Fcafd-lychee for lychee picking points, which is able to utilize the depth camera point cloud stream to directly detect the lychee picking points and get their 3D coordinates, thus accelerating and efficiently carrying out the picking point detection.

\section{Materials and Methods} \label{section: method}
\subsection{System overview}
\begin{figure}[htbp]
\centering
\includegraphics[width=8.5cm]{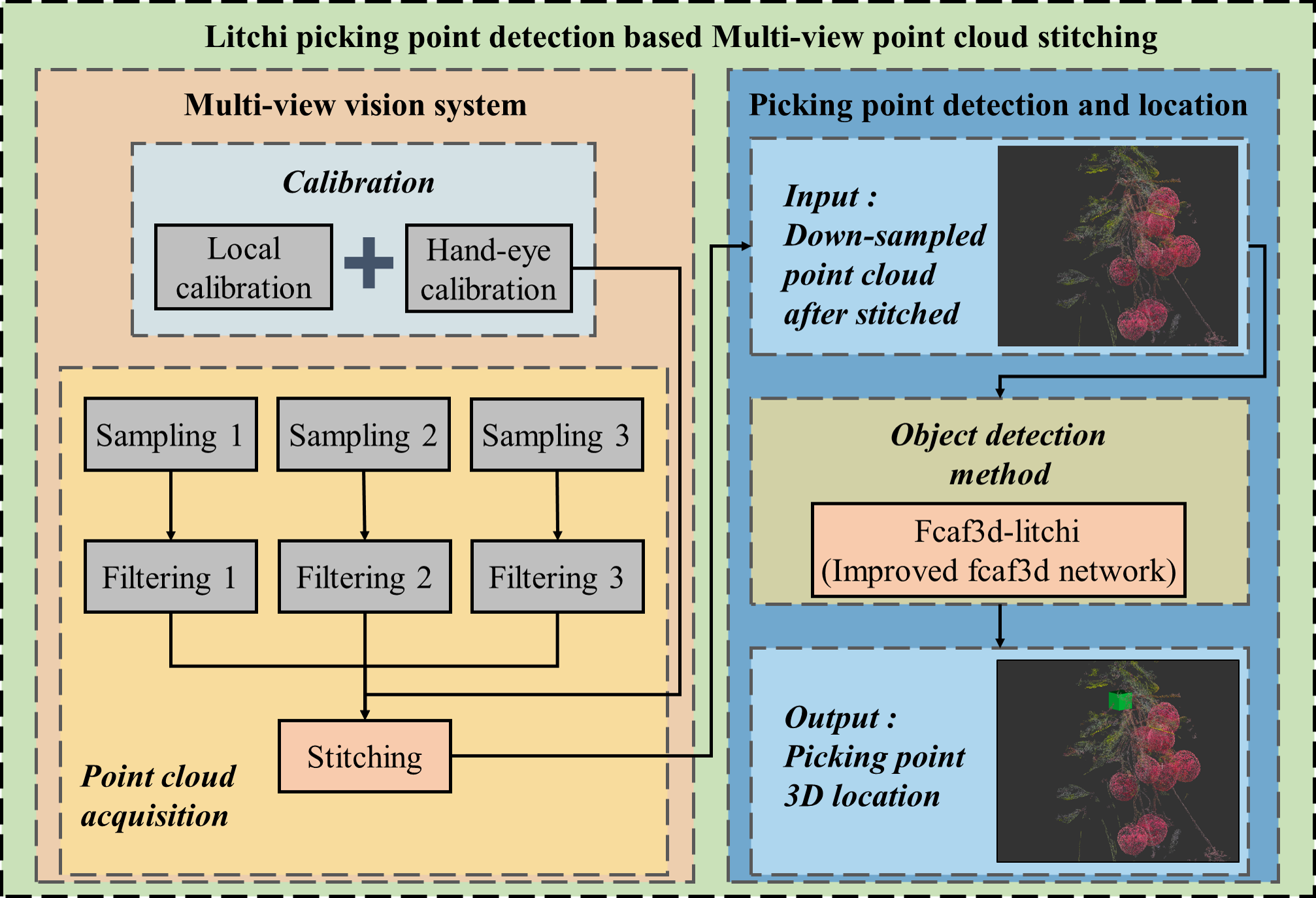}
\caption{Lychee picking point robot system and methods.} 
\label{fig:graph1}
\end{figure}

The proposed lychee picking point detection process is shown in Fig. \ref{fig:graph1}. The process can be divided into two steps, acquiring point cloud and performing multi-view stitching, and Fcaf3d-lychee modeling for precise location. First, the picking robot, which has carried the orchard map information, autonomously walks in the orchard and will obtain the initial position of the target from the point cloud data stream. After that, the picking robot will perform the acquisition of the set three-angle point cloud at a close distance to the target and perform stitching and downsampling. The pre-processed point cloud will be fed into the Fcaf3d-lychee model for accurate identification and localization. Finally, the end-effector will be guided to perform the lychee picking action.

\subsection{Hand-Eye vision model \& Closed-Loop calibration method}
The camera sensor was operated on the aubo robotic arm with six degrees of freedom and collected data, which ultimately transformed the 3D coordinates of the lychee picking point accurately to the coordinate system of the robotic arm base. Prior to data acquisition, camera calibration and hand-eye calibration were completed respectively. The former is realized in this study by the classical Zhang's calibration method proposed by \cite{r59}. In our early work some research has been done on the calibration between the camera and the robotic arm, and \cite{r60} proposed a robust and accurate closed-loop hand-eye calibration method to determine the coordinate correlation between the end flange of the robotic arm and the camera sensor for a robot with an eye in the hand. The Hand-eye matrix as eq. (1).
\begin{gather} 
    \widehat{{_{C}^{F}}T} =\frac{1}{N_c} \sum_{i=1}^{N_C} {({_{B}^{C}}T{^{(i)}}{_{R}^{B}}T{_{F}^{R}}T{^{(i)}})^{-1}}
\end{gather}
In which, the $_R^BT$ is the fixed transformation matrix \{R\} with respect to \{B\} since the relative fixed between the base of the robot arm and the checkerboard during the calibration process, which can be calibrated by controlled proximity of the origin of \{G\} to the origin as well as the X and Y axes of \{B\}. the $_B^CT^{(i)}$ is the transformation matrix \{B\} with respect to \{C\}, corresponding to the i th position of the robot that can be obtained by the least squares method \cite{r61}. $_F^RT^{(i)}$ is the transformation matrix {F} with respect to {R}, corresponding to i th position of the robot that can be easily obtained by the forward kinematics. The $N{_c}$ is the number of different poses. In this study, $N{_c}$=16. 

As the robotic arm moved at certain positions \cite{r64}, the local features at each location of the lychee picking point were eventually transformed into the same base coordinate system to make up the global feature (though it should still be called “local feature” because the arm with the camera sensor was limited to a localized area when scanning).

\subsection{Point cloud acquisition}
\subsubsection{Filtering}
Many factors may influence the accuracy of raw point clouds obtained multi-view vision system, such as changes in natural illumination, vibration in dynamic environments, hand-eye calibration error, and the camera’s inherent hardware error. These factors generate messy noise and uneven structures among the acquired point cloud. As discussed in this section, we combined a statistical filter and color filter to clear the outliers scattered outside the main body and with wide color gap from the picking point to provide a good initial state for subsequent stitching.

As proposed by \cite{r62} for filtering for color in 3D reconstruction of fruit trees, we propose to perform simple color filters in point clouds with color features to initially reduce a large number of green leaves and provide good initial conditions for subsequent point cloud detection.
\begin{gather} 
\left\{
\begin{aligned}
    R_s & > & \sigma_1 \\
    G_s & \leq & \sigma_2
\end{aligned}
\right.
\end{gather}
where, $R_s$, $G_s$ are the red and green channels in the color features of the point cloud respectively, when $R_s\leq\sigma_1$, $G_s$\textgreater$\sigma_2$, the point cloud will be removed out of the whole point cloud.

The statistical filter can effectively remove any anomalous points that are scattered outside of the main structure of the point cloud \cite{r58}. For arbitrary elements, all field points of each point are searched and the distance from each point to its neighbor calculated. The mean $\mu$ and the standard deviation $\sigma$ of the distances of all points from their neighbors are calculated, and any point whose mean value is outside of the standard interval will be considered as noise.
\begin{gather} 
\left[ 
    \mu-\alpha_v\times\sigma,\ \mu+\alpha_v\times\sigma
\right]
\end{gather}
where $\alpha_v$\textless3 determines the width of the standard interval.

\begin{figure}[H]
\centering
\includegraphics[width=8.5cm]{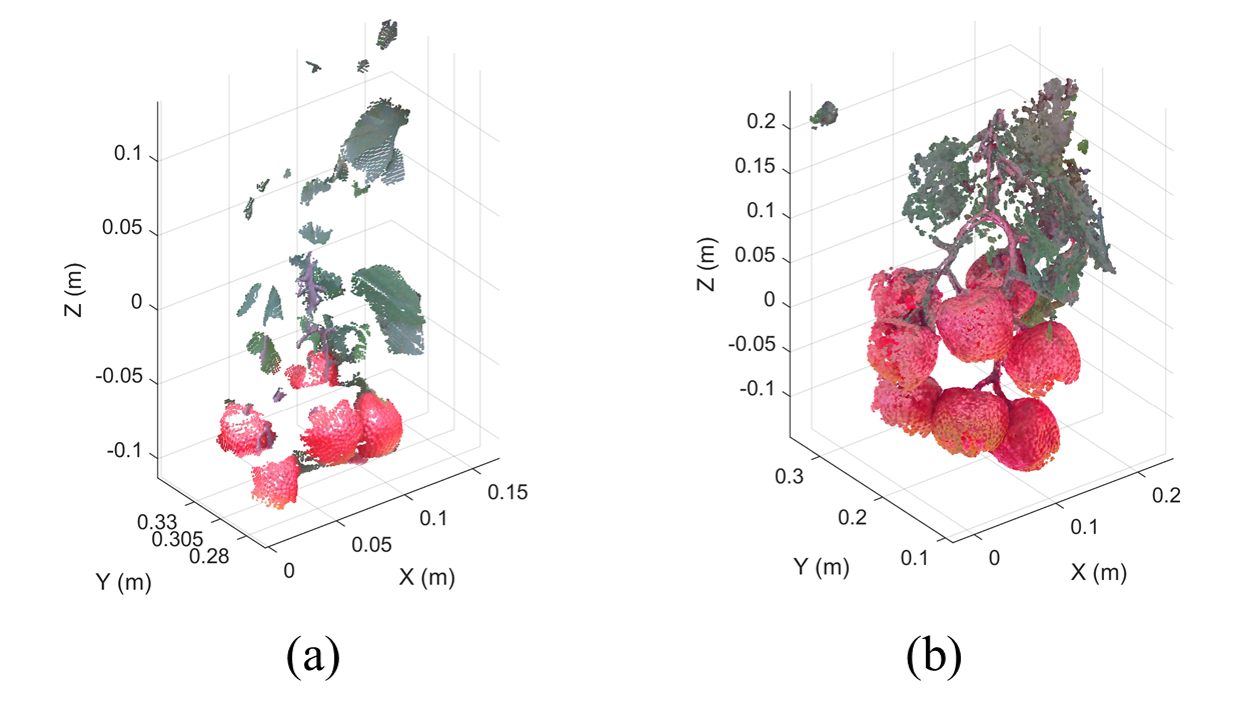}
\caption{After color and statistical filter.} 
\label{fig:graph2}
\end{figure}

The filtered point cloud is shown in Fig. \ref{fig:graph2} (a).

\subsubsection{Stitching}
According to the hand-eye calibration theory, the orientation of each group of point clouds (three specific angles) in the coordinate system of the robotic arm base can be obtained, which provides a good initial orientation for the following point cloud stitching.

Suppose the number of point cloud points obtained by the camera at specific angles for $n_1$, $n_2$ and $n_3$ respectively. The coordinates of the k th point relative to the base coordinate system are ${^R}P_A^{(k)}$, ${^R}P_B^{(k)}$ and ${^R}P_C^{(k)}$ (A, B, C only represent different positions of the camera) those to the camera coordinate system are ${^{CA}}P^{(k)}$, ${^{CB}}P^{(k)}$ and ${^{CC}}P^{(k)}$ (CA is the representation of the camera coordinate system at position A, CB, CC are similar), respectively. Substitute all the values of k to obtain the points relative to the base coordinate system:
\begin{equation}
\left[
\begin{array}{c}
    {^R}P_A^{(1)} \\
    {^R}P_A^{(2)} \\
    \vdots \\
    {^R}P_A^{(n_1)} 
\end{array}
\right]
= \ 
^{\ R}_{CA}T
\left[
\begin{array}{c}
    ^{CA}P^{(1)} \\
    ^{CA}P^{(2)} \\
    \vdots \\
    ^{CA}P^{(n_1)} 
\end{array}
\right]
\end{equation}

\begin{equation}
\left[
\begin{array}{c}
    {^R}P_B^{(1)} \\
    {^R}P_B^{(2)} \\
    \vdots \\
    {^R}P_B^{(n_2)} 
\end{array}
\right]
= \ 
^{\ R}_{CB}T
\left[
\begin{array}{c}
    ^{CB}P^{(1)} \\
    ^{CB}P^{(2)} \\
    \vdots \\
    ^{CB}P^{(n_2)} 
\end{array}
\right]
\end{equation}

\begin{equation}
\left[
\begin{array}{c}
    {^R}P_C^{(1)} \\
    {^R}P_C^{(2)} \\
    \vdots \\
    {^R}P_C^{(n_3)} 
\end{array}
\right]
= \ 
^{\ R}_{CC}T
\left[
\begin{array}{c}
    ^{CC}P^{(1)} \\
    ^{CC}P^{(2)} \\
    \vdots \\
    ^{CC}P^{(n_3)} 
\end{array}
\right]
\end{equation}
In which, $_{CA}^{\ R}T$= $_{CB}^{\ R}T$=$_{CC}^{\ R}T$=$_{C}^{R}T$=$_{F}^{R}T^{(i)}$$\widehat{{_{C}^{F}}T}$, is the transformation matrix between the camera coordinate system and the base coordinate system of the robotic arm deduced in Sec. III-B. Thus, the stitched point cloud $P_s$ can be written as:
\begin{equation}
\setlength{\arraycolsep}{1.8pt}
    P_s
=
\left[
\begin{array}{c|c|c}
    \left[
    \begin{array}{c} 
    {^R}P_A^{(1)} \\
    {^R}P_A^{(2)} \\
    \vdots \\
    {^R}P_A^{(n_1)} 
    \end{array}
    \right]^T
    \!& 
    \left[
    \begin{array}{c} 
    {^R}P_B^{(1)} \\
    {^R}P_B^{(2)} \\
    \vdots \\
    {^R}P_B^{(n_3)} 
    \end{array}
    \right]^T
    & 
    \left[
    \begin{array}{c} 
    {^R}P_C^{(1)} \\
    {^R}P_C^{(2)} \\
    \vdots \\
    {^R}P_C^{(n_3)} 
    \end{array} 
    \right]^T
\end{array}
\right]^T
\end{equation}

The effect of point cloud stitching is shown in Fig. \ref{fig:graph2} (b).

\begin{figure*}[htbp]
\centering
\includegraphics[width=\textwidth]{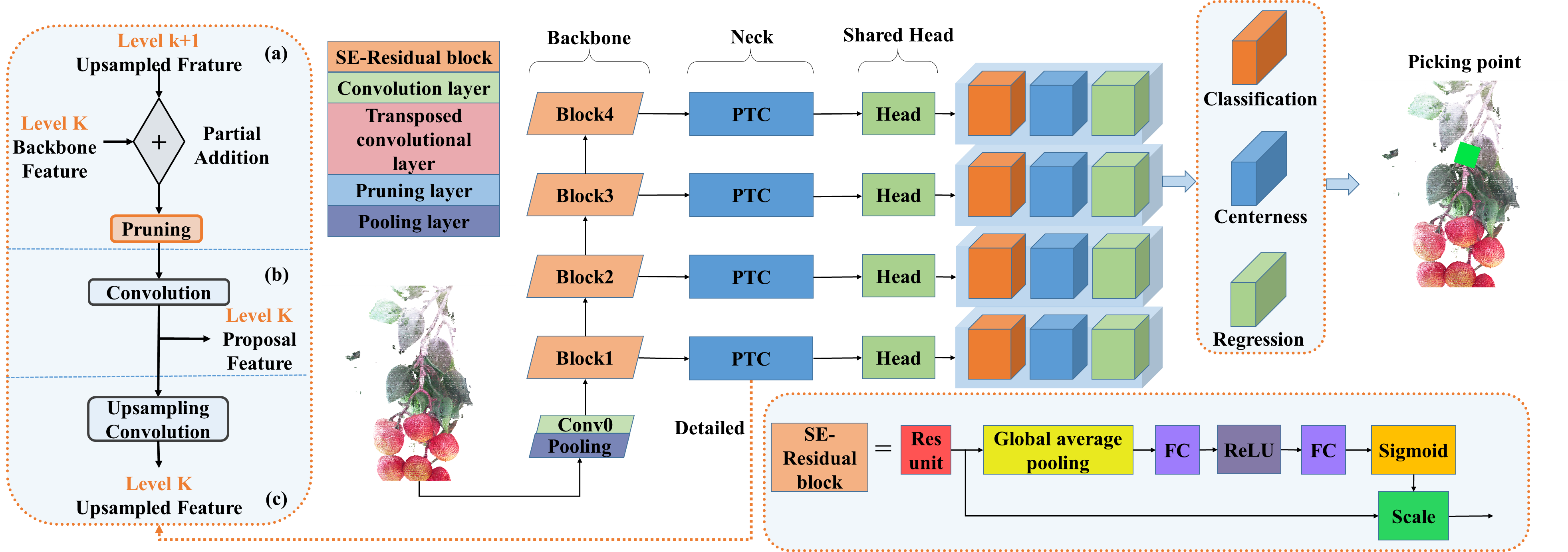}
\caption{Fcaf3d network structure.} 
\label{fig:graph3}
\end{figure*}

\subsection{Fcaf3d-lychee for lychee picking point detection}
\cite{r37} proposed 3D target detection network only using point cloud data as material, which known as Fcaf3d (Fully Convolutional Anchor-Free 3D Object Detection). It is a network model that greatly improves detection generalization specifically designed to achieve fast and accurate detection of indoor 3D objects. This provides technical support for the identification and localization of lychee picking points. Fcaf3d network algorithm is an improved algorithm based on \cite{r38} proposed GSDN (Generative sparse detection networks). Among the improvements, Fcaf3d proposed a fully convolutional anchorless box and a novel OBB (oriented bounding box) parametrization, which improves mAP@.5 by 11.4\% in comparison with GSDN. Thus, the Fcaf3d network algorithm provides the best method for the identification and localization of lychees picking points.

There are four main components of the Fcaf3d network model: Input, Backbone, Neck, and Head. We modified its structure and combined the SE module with every Res unit in the backbone to form the SE-Res Block. In this paper, we will call it Fcaf3d-lychee network model. The structure of the Fcaf3d-lychee model is shown in Fig. \ref{fig:graph3}.

The sparseness of the point cloud and the fact that the observable surface of the point cloud does not intersect with the center of the instance, in addition to the problem of high computational cost of 3d convolution, cause great difficulties in the task of 3D object detection. Instead of normal dense 3d convolution computation, Fcaf3d and GSDN both adopted the computation proposed by \cite{r39} for sparse convolution of point cloud. The calculation is as in eq. (8).

\noindent Let $\textit{x}_\textbf{u}^{in} \in \mathbb{R}^{N^{in}}$ be an $N^{in}$-dimensional input feature vector in a D-dimensional space at $\mathbf{u} \in \mathbb{R}^D$ (a D-dimensional coordinate), and convolution kernel weights be $\textbf{W} \in \mathbb{R}^{{K^D}\times{N^{in}}\times{N^{out}}}$. Next, break down the weights into spatial weights with $K^D$ matrices of size ${N^{in}}\times{N^{out}}$ as $\textit{W}_\textbf{i}$ for $|\{\textbf{i}\}_\textbf{i}|=K^D$. Then the generalized sparse convolution is as follows:
\begin{equation}
\textbf{x}^{out}_\textbf{u}
= \ 
\sum_{i \in \mathbb{N}^D (\textit{u},\mathbb{C}^{in})}^{}W_\textbf{i}\textbf{x}_{\textbf{u}+\textbf{i}}^{in} \ \text{for} \ \textbf{u} \in \mathbb{C}^{out}
\end{equation}
where $\mathbb{N}^D$ is a set of offsets that define the shape of a kernel and $\mathbb{N}^D (\textbf{u},\mathbb{C}^{in})=\{\textbf{i}|\textbf{u+i}\in C^{in},\textbf{i}\in \mathbb{N}^D\}$ as the set of offsets from the current center, \textbf{u}, that exist in $\mathbb{C}^{in}$. $\mathbb{C}^{in}$ and $\mathbb{C}^{out}$ are predefined input and output coordinates of sparse tensors.

The sparse tensor is different from previous ones in that it is batch-indexed, i.e., new sparse tensor coordinates are given an index number to distinguish points from different batches occupying the same coordinates, and it is in COO format to enable efficient querying of neighbor coordinates. Fcaf3d’s Backbone adopts a variant form of the same high-dimensional residual structure used in GSDN, which consists of a series of step-by-step convolutions as well as residual blocks, each of which reduces the resolution of the target space and exponentially increases the size of the receptive field. In the Neck, a transposed convolution is used to improve the features of the large low-resolution receptive field generated by the previous residual block, and jump-joining with sparse pruning is utilized to control the sparse tensor of the cubic growth. The Head, on the other hand, consists of parallel sparse 3d convolutional layers that output the classification probability, bounding box regression parameters, and bounding box centrality of the layer, respectively. On this basis, in order to improve the accuracy and generalization, Danila Rukhovich et al. proposed the anchor-box-free proposal and the fuzzified bounding box parameters using the idea of Möbius rings, which greatly reduces the number of hyper-parameters during training, and the generalization and accuracy are also dramatically improved.

However, the result of Fcaf3d is undesirable for the identification and localization of lychee picking points with small targets. Therefore, this paper proposes the Fcaf3d-lychee model for 3d point cloud detection of picking points of lychee. For small targets, especially small targets in the point cloud, the information in the point cloud is inhomogeneous, and the features of the small targets may be weak with respect to the features in the whole point cloud, which is very important for the detection of small targets. Additionally, in the residual structure, the number of input channels per residual block is too large, while a small number of channels is better for retaining some special information about small targets in deep features. In conclusion, the following improvements are made in this paper for the identification and localization of lychee picking points: (1) Se-block: Combined the se-attention mechanism module with each res unit in the backbone, resulting in the SE-Residual module; (2) Hyperparameters: For the four residual blocks of the resnet, the original input channels are (64, 128, 256, 512), and we get the optimal number of channels for the experiments to be (64, 128, 256, 256), which not only improves the accuracy of the model but also can reduce the occupation of the video memory greatly.

\begin{figure*}[htbp]
\centering
\includegraphics[width=\textwidth]{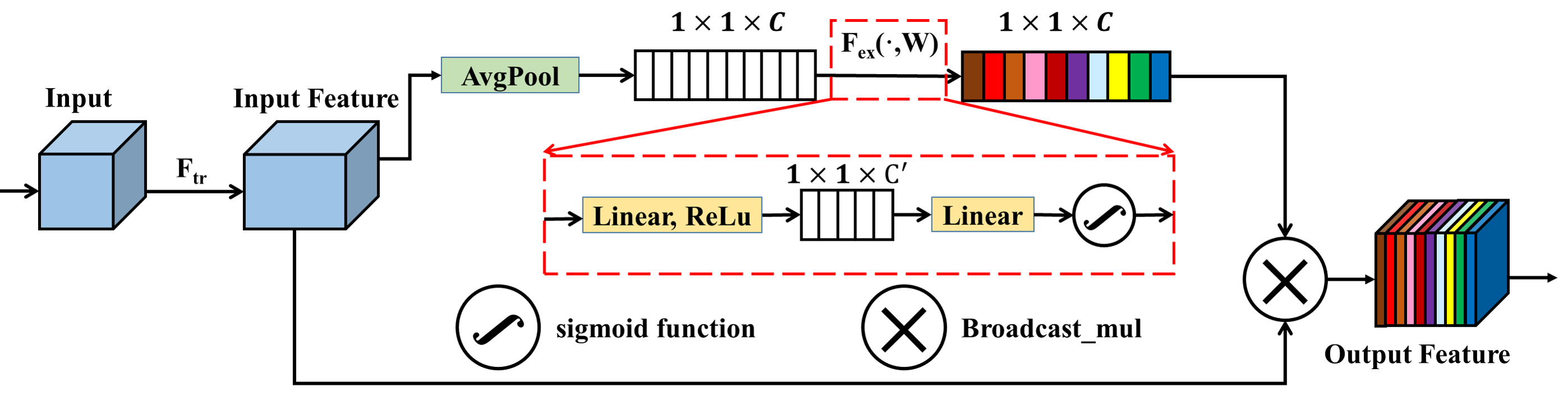}
\caption{Structure of SE module.} 
\label{fig:graph4}
\end{figure*}

\subsection{Squeeze-and-Excitation (SE) module}
Soft attention mechanism generally includes channel attention mechanism and spatial attention mechanism. The spatial attention module focuses on the spatial location information of the data, while the channel attention module pays more attention to the channel information of the features. SENet belongs to channel attention, which weights the channels of different features to establish the relationship between the channels of the features, and the introduction of SENet will not change the original spatial dimensions. This is a good choice for sparse and non-uniform data such as point cloud, it just automatically obtains the weights of the feature channels and then uses the obtained weights to filter the key features. 

The SE module consists of two stages: Squeeze and Excitation. In the Squeeze stage, the original feature map is dimensionally reduced through global average pooling, resulting in a single value for each channel. This value represents the global importance of each channel. In the Excitation stage, a fully connected layer is used with an input dimension equal to the number of channels and an output dimension significantly smaller than the number of channels. The output is then processed through the ReLU activation function to learn the weights for each channel. These weights are employed to reweight the features of each channel \cite{r40}. The Squeeze-and-Excitation attention mechanism was expressed as:
\begin{equation}
\begin{split}
\textbf{$M_c$}(F)
&=F\cdot\sigma(L_2(ReLu(L_1(AvgPool(F)))))\\
&=F\cdot\sigma(L_2(ReLu(L_1(F_{avg}))))
\end{split}
\end{equation}
where \textbf{$M_c$} is the attention map, the input feature map is F, $\sigma$ represents the sigmoid functions, ReLu is the Non-linear, $L_1$ and $L_2$ are the full connection operation. $F_{avg}$ is the average pooling operations on the feature map. The SE attention module is shown in Fig. \ref{fig:graph4}.

\subsection{Lychee Picking Point Dataset}
\subsubsection{Data collection}
This study was conducted from June 23 to 24, 2023, at lychee Expo Park, East Huanshi Road, Jiangpu Street, Conghua District, Guangzhou City, Guangdong Province, China (Longitude $23^\circ59^\prime$E, Latitude $113^\circ62^\prime$N). A total of 850 groups (three point cloud per group, in preparation for subsequent stitching) of point cloud data were collected from the dwarf lychee species "Huai Zhi" using the depth camera Azure Kinetic DK. These lychee point cloud data will be used to build a lychee identification and detection dataset. Some invalid point cloud data with severe occlusion overlap and extremely dark point cloud data were discarded, leaving 800 point cloud data obtained from different shooting distances as well as different lighting conditions.

\subsubsection{Voxel Downsampling}
\begin{figure}[htbp]
\centering
\includegraphics[width=0.5\textwidth]{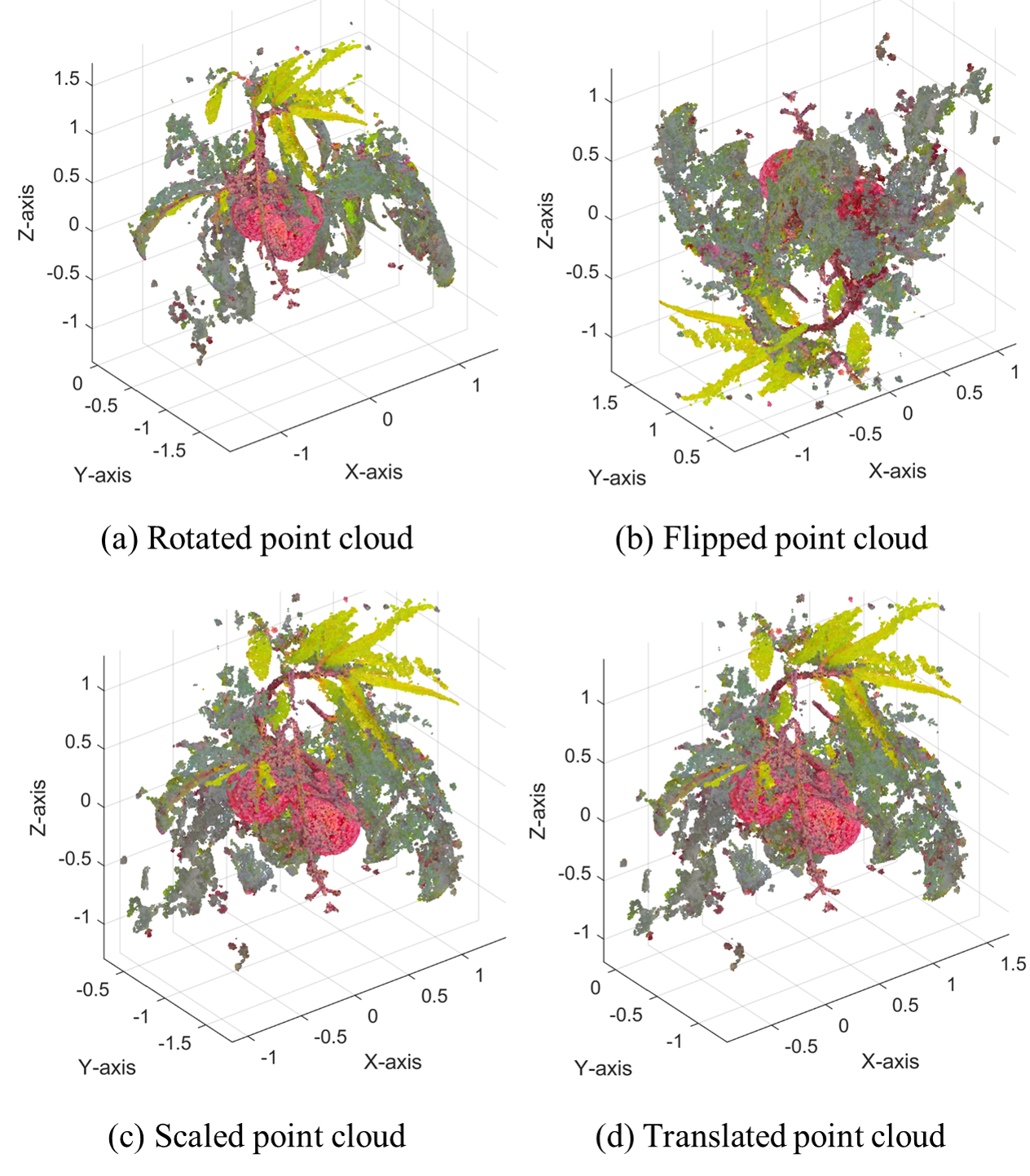}
\caption{Point cloud augmentation.} 
\label{fig:graph5}
\end{figure}

As shown in Fig. \ref{fig:graph5}, in this paper, a series of data enhancement operations are carried out on the original dataset, and data preprocessing is a very important process, especially for sparse data such as point cloud, the effect of preprocessing has a great impact on the subsequent production of datasets and model training, and the preprocessing of the dataset prior to the training of the model will be described in this subsection. Preprocessing includes but is not limited to, overall rotation, translation, scaling, and flipping of the data, but the most important pre-processing steps are multi-view point cloud alignment and voxel downsampling. These preprocessing operations are very important in the training of the model, especially for the special data of the point cloud, these operations improve the robustness of the data, greatly improve the training accuracy of the model, minimize the overfitting phenomenon of the model, and improve the generalization ability of the model training. 

Although the above pre-processing does not reduce the training difficulty of the model, the voxel representation is used to further compress the point cloud data without destroying the original information. In this study, point cloud downsampling is used in open3d point cloud filtering. The algorithm is to create a uniformly downsampled point cloud from the input point cloud using a regular voxel grid in a two-part operation, where first the input point cloud is voxelised, followed by the centre of mass of the point cloud within the voxel de-voxelisation of all non-empty voxels as the location of the point of the final modified voxel. In this way, we further reduce the number of point clouds, greatly reducing the performance requirements, and also line to improve the training speed and accuracy, but also improve the final generalisation of the model. For the choice of voxel size, this study set the voxel size to 0.01m after many trials and experiments, from the initial two hundred thousand or so points, reduced to less than one hundred thousand points, and more importantly, this is the training of the model has a very obvious increase in the accuracy of training, but also greatly improved the generalisation of the model. Specific training results are given in the experimental part of section \ref{section: experiment}.

\section{Experiment and Discussion} \label{section: experiment}
\subsection{Dataset and Experimental Setup}
In this paper, the specific configuration of the experimental setup is organized in Table \ref{tab: Model training environment settings}. The Fcaf3d-litchi model was trained on the mmdetection3d framework. The specific steps of training are as follows: as in Table \ref{tab: lychee picking points dataset split used for training}, the dataset is divided into a training set and a validation set according to the ratio of 8:2, thus the model is trained on 800 point cloud data divided into small batches, and the network model adopts the Adamw optimizer, with the initial learning rate set to 0.001 and the weight decay set to 0.0001. The same training parameters and platform were also used to train Votenet, Tr3d, and the initial Fcaf3d models for comparison.
\begin{table}[htbp]
\centering
\caption{Model training environment settings.}
\label{tab: Model training environment settings}
\scalebox{0.95}{
\begin{tabular}{cc}
\hline 
{Parameter} & {Configuration} \\
\hline 
Operating system & Ubuntu 20.04 \\
Deep learning framework & 1.8.0+cu113 \\
Programming language & Python 3.8 \\
CUDA & 11.3 \\
CPU & Intel(R) Core(TM) i5-9400F CPU@2.90 GHz \\
GPU & NVIDIA GeForce RTX 2070 super \\
\hline 
\end{tabular}
}
\end{table}

\begin{table}[htbp]
\centering
\caption{Lychee picking points dataset split used for training.}
\label{tab: lychee picking points dataset split used for training}
\begin{tabular}{ccc}
\hline 
{ } & {Point cloud} & {Picking point ground truth} \\
\hline 
Train & 640 & 1964 \\
Validation & 160 & 516 \\
\hline 
\end{tabular}
\end{table}

\subsection{Metrics}
For the evaluation of model training, this study will use, but not be limited to, metrics such as Precision, Recall, and $F_{1}$ score to assess the performance of the trained model on the test dataset. The use of the $F_{1}$ score is motivated by its combination of Recall (R) and Precision (P), making it a comprehensive metric for performance evaluation. Therefore, many previous studies on fruit detection have widely adopted the $F_{1}$ score as one of the key evaluation metrics. Accuracy is an important reference parameter to determine whether a model is performing well or not. TurePositive(TP) is the number of lychee picking points identified by the model, FalsePositive(FP) is the number of lychee picking points over-identified by the model, FalseNegative(FN) is the number of lychee picking points under-identified by the model.The expressions for Precision, Recall, $F_{1}$ score and Accuracy are as follows:
\begin{gather}
   Precision = \frac{TP}{TP+FP}
   \\
   Recall = \frac{TP}{TP+FN}
   \\
   F_1 = \frac{2\times Precision\times Recall}{Precision+Recall}
   \\
   Accuracy = \frac{TP+TN}{TP+FN+FP+TN}
\end{gather}

\subsection{Comparsion of model training results}
To verify the recognition and detection performance of the improved Fcaf3d-lychee model, this study compares it with three classic 3D detection neural networks: Fcaf3d, Votenet, and Tr3d. Additionally, Fcaf3d-lychee with different voxel sizes for preprocessing is also compared. Table \ref{tab: Test results of the 4 models} presents the results of these models and the improved Fcaf3d-lychee with varying voxel sizes for preprocessing.

\begin{figure}[t]
\centering
\includegraphics[width=0.5\textwidth]{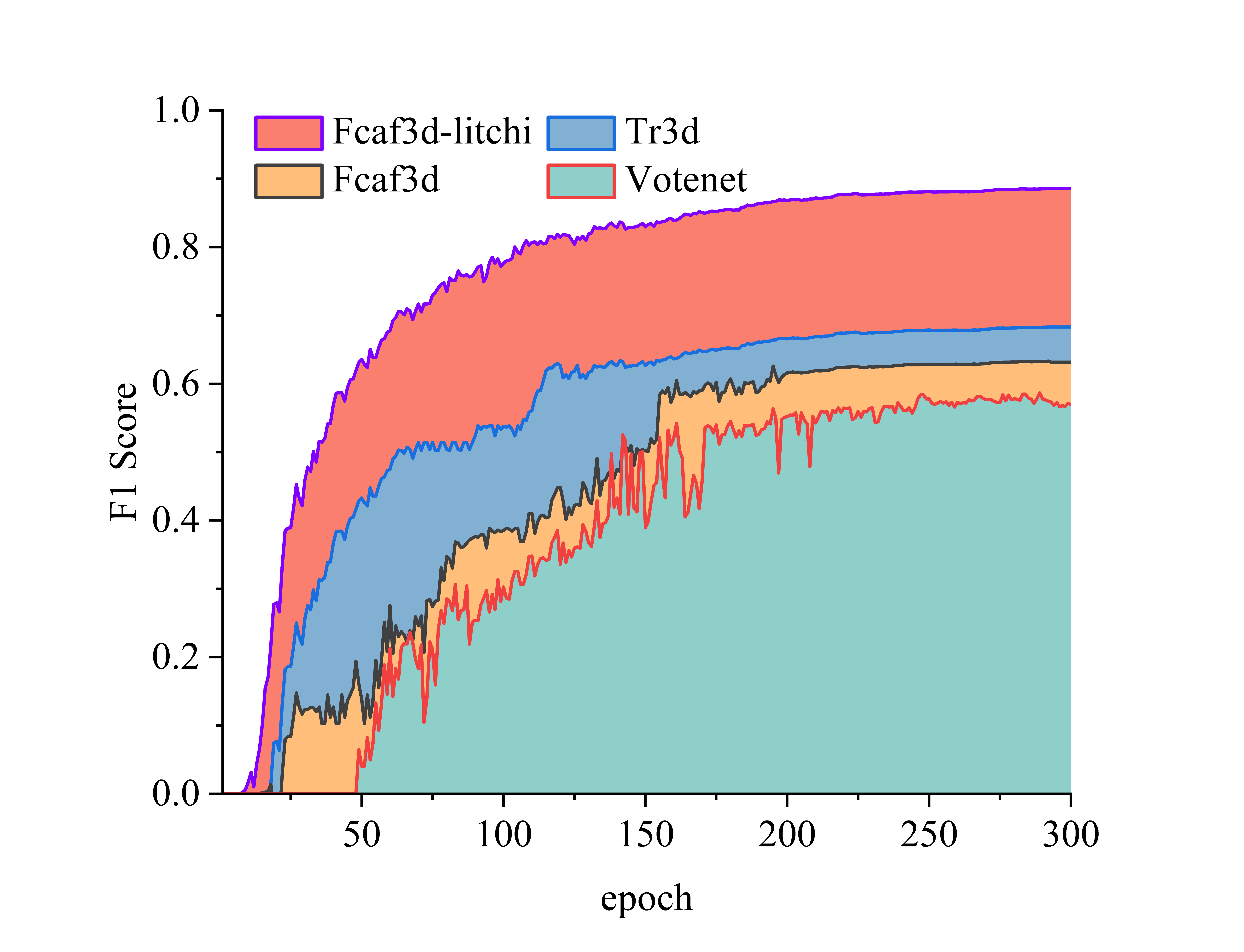}
\caption{$F_{1}$ score achieved by four models Fcaf3d, Votenet, Tr3d, and Fcaf3d-lychee model after 300 epochs training.} 
\label{fig:graph6}
\end{figure}

\begin{table}[h]
\centering
\caption{Test results of the 4 models.}
\label{tab: Test results of the 4 models}
\begin{tabular}{cccc}
\hline 
{Model name} & {Precision(\%)} & {Recall(\%)} & {$F_{1}$ score(\%)} \\
\hline 
Fcaf3d & 60.03 & 66.67 & 63.18 \\
Votenet & 53.31 & 61.11 & 56.94 \\
Tr3d & 63.12 & 74.44 & 68.31 \\
Fcaf3d-lychee & 83.39 & 94.44 & 88.57 \\
\hline 
\end{tabular}
\end{table}

As shown in Table \ref{tab: Test results of the 4 models}, using the same dataset and voxel preprocessing sizes, the results of the proposed Fcaf3d-lychee model are noticeable compared to other detection models. In comparison to Fcaf3d, the Fcaf3d-lychee model exhibits a 23.36\% improvement in Precision, a 27.77\% increase in Recall, and a 25.36\% rise in $F_{1}$ score. When compared to Votenet, Fcaf3d-lychee shows a 30.08\% improvement in Precision, a 33.33\% increase in Recall, and a 31.63\% rise in $F_{1}$ score. Compared to Tr3d, the Fcaf3d-lychee model demonstrates a 20.27\% increase in Precision, a 20.00\% rise in Recall, and a 20.26\% improvement in $F_{1}$ score. The $F_{1}$ score curves for each model are depicted in Fig. \ref{fig:graph6}, showing significant improvement in the $F_{1}$ score curve of the proposed model compared to other models.

\subsection{Ablation Study}
To further validate the improvements proposed in this paper for the Fcaf3d-lychee model, this section conducts ablation experiments to examine the performance variations and provide a comprehensive understanding of the contributions and impact of each improvement on the model's performance. We experiment with different voxel sizes, varying input channel configurations in residual blocks, and comparing models with and without SE attention modules. The results of the ablation experiments are summarized in Table \ref{tab: Ablation Study}.

\subsubsection{Voxel Size Preprocessing} 
The experiments involve using different voxel sizes for preprocessing. The results indicate that, for the dataset used in this paper, the optimal voxel size for preprocessing is a voxel cubic grid of 0.01m. Larger voxel sizes lead to a significant loss of information for small objects in the point cloud, resulting in a rapid decline in detection performance. On the other hand, excessively small preprocessing voxel sizes make the overall structure overly complex, hindering effective detection.

\subsubsection{SE attention}
With optimized voxel preprocessing, experiments are conducted with and without attention mechanisms. The results show a significant improvement in detection performance with the introduction of the SE attention mechanism. The channel attention mechanism with a compression ratio of 16 enhances the detection performance of the proposed Fcaf3d-lychee model.

\subsubsection{Res in-channel} 
After incorporating the SE attention mechanism, modifications are made to the input channels in residual blocks. The results from the summarized table demonstrate that adjusting the channel numbers to (64, 128, 256, 256) significantly improves the model's performance. The model achieves its best performance with an accuracy of 77.84\% and a recall rate of 94.44\%. Optimizing the channel numbers effectively enhances the model's performance while maintaining stability and robustness.

\begin{table*}
\caption{Ablation Study on Voxel Size, SE-Attention, and Residual Block Input Channels in Fcaf3d-lychee. The better options are marked in bold.Values represent the percentage (\%) of Precision, Recall, and $F_{1}$ Score.}
\label{tab: Ablation Study}
\centering
\begin{tabular}{ccccc}
\hline 
    {Ablating parameter} & {Voxel size(m)} & {Precision} & {Recall} & {$F_{1}$ score} \\
    \hline 
    \multirow{2}*{Voxel size(m)}&NO&60.03&66.67&63.18\\
    &{\bf{0.01}}&{\bf{70.65}}&{\bf{88.89}}&{\bf{79.16}}\\
    \cline{2-5}
    \multirow{2}*{SE attention}&NO&70.65&88.89&79.16\\
    & {\bf{YES}} &{\bf{75.52}}&{\bf{94.44}}&{\bf{83.92}}\\
    \cline{2-5}
    \multirow{2}*{Res\_inchannel}&(64, 128, 256, 512)&75.52&94.44&83.92\\
    &{\bf{(64, 128, 256, 256)}}&{\bf{83.39}}&{\bf{94.44}}&{\bf{88.57}}\\
\hline 
\end{tabular}
\end{table*}

\subsection{Visual Sensing in Orchards}
\begin{figure}[ht]
\centering
\includegraphics[width=8.5cm]{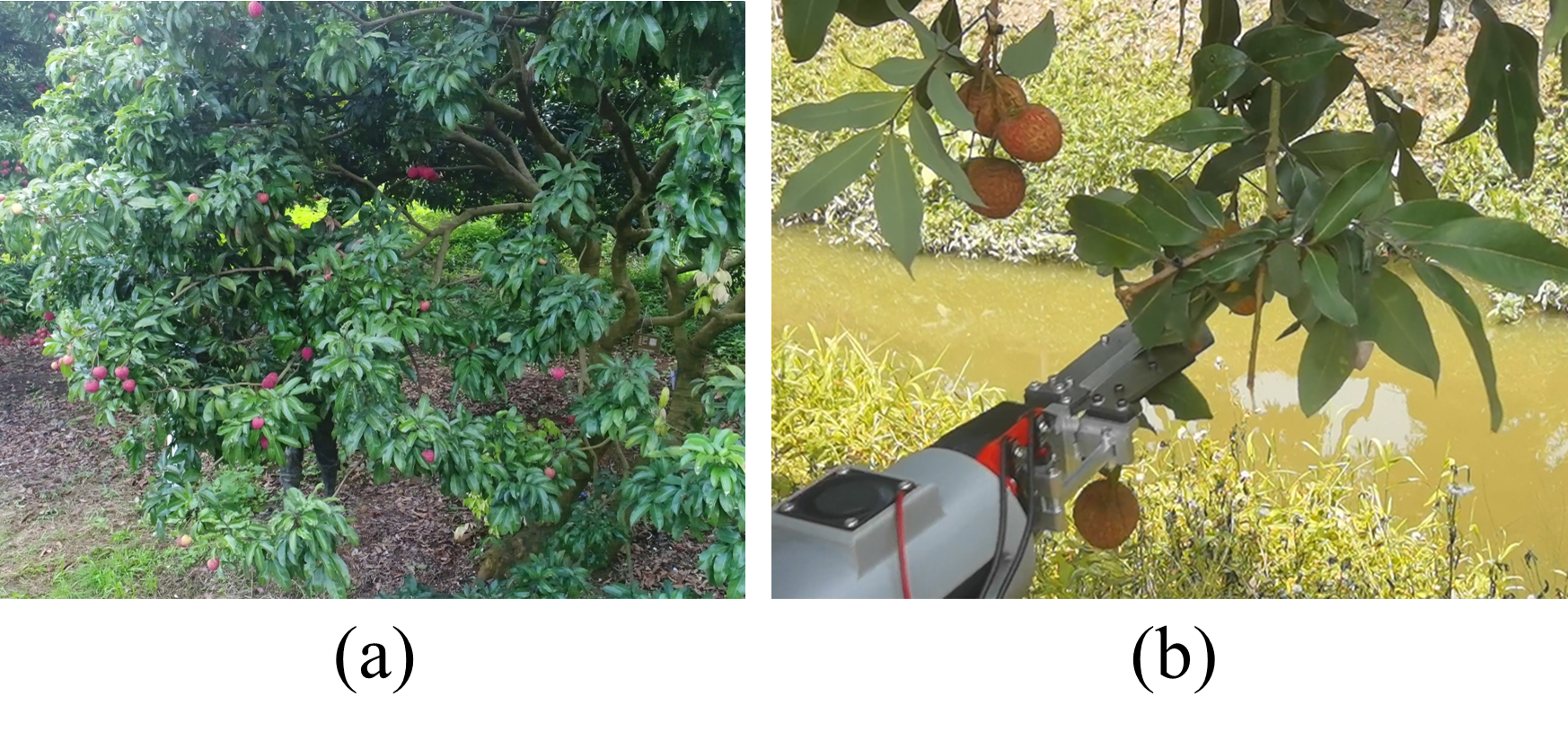}
\caption{Lychee picking robot equipped with Fcaf3d-lychee model in (a) Guangzhou lychee Expo Park, (b) The success rate of picking point localization is evaluated by performing cutting action by the end-effector of the picking robot.} 
\label{fig:graph7}
\end{figure}

In this section, we show the results of 3D inspection in a lychee garden. This experiment was conducted on July 2, 2023, from 1-3 pm. The orchard environment road surface is not very rugged. The data was collected in the afternoon under strong sunlight. In our previous work, \cite{r67} proposed a new semantic mapping and navigation framework for autonomous navigation of orchard robots. The picking robot moved on the aisle, and the Fcaf3d-lychee model acquired the fruit of the target lychee in the real-time point cloud stream as the robot performed further precise positioning of the picking point location (As shown in Fig. \ref{fig:graph7}). When the position of the target lychee is obtained, the robotic arm will be a depth camera suitable depth for three directions of point cloud acquisition and point cloud stitching in Sec. 3.3, and then the detection results will be obtained in the Fcaf3d-lychee model. 

\begin{table}[htbp]
\centering
\caption{Performance with different occlusion situations by Fcaf3d-lychee.}
\label{tab: Performance}
\begin{tabular}{cccc}
\hline 
{\bf{Occlusion situation of picking points}} & {\bf{TP}} & {\bf{FP}} & {\bf{Accuracy}} \\
\hline 
Unobstructed&232&17&0.932\\
Slight obstructed&103&22&0.824\\
Severely obstructed&75&23&0.765\\
\hline 
\end{tabular}
\end{table}

\begin{figure*}[htbp]
\centering
\includegraphics[width=\textwidth]{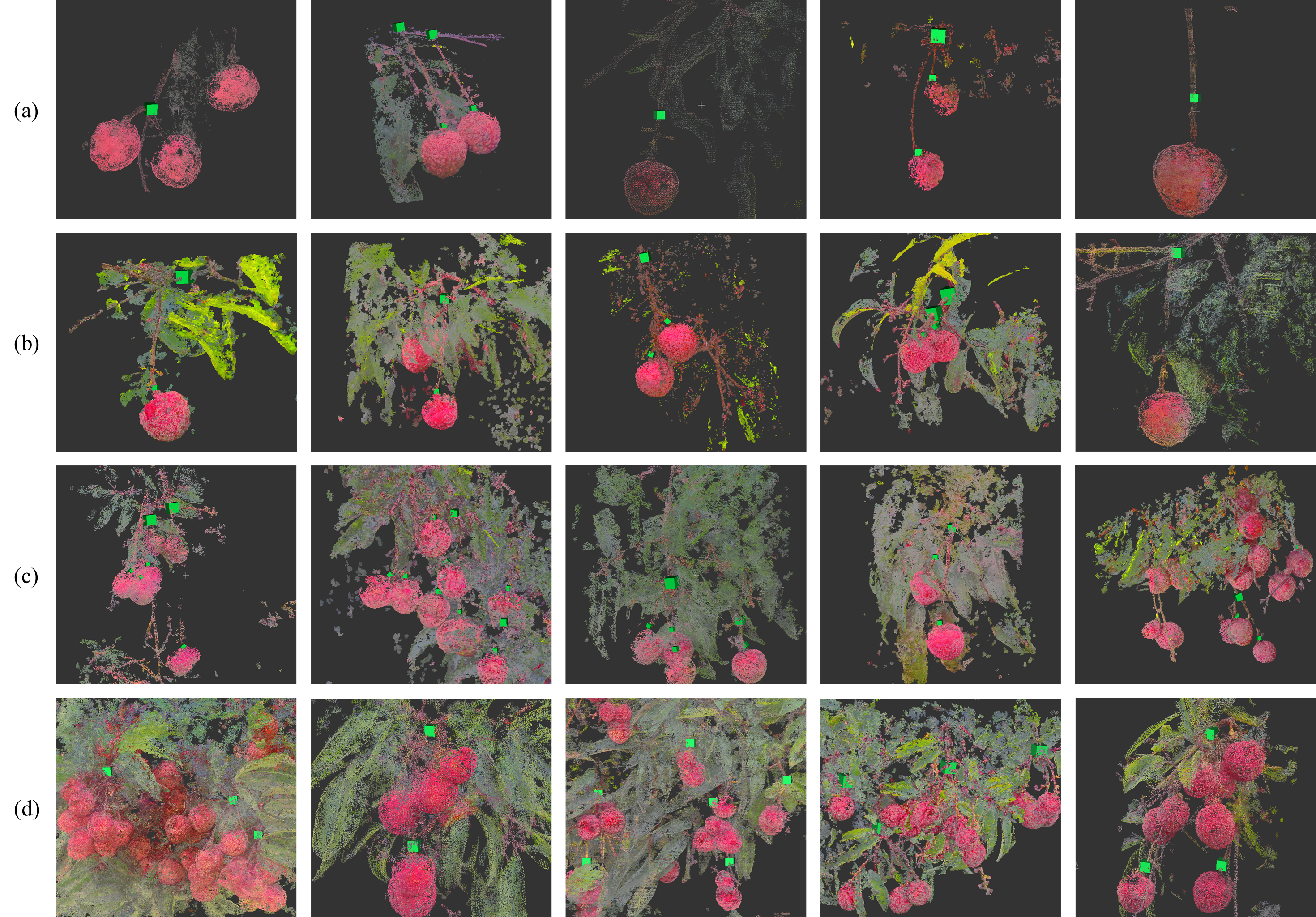}
\caption{Detection results in the lychee orchard by using the Fcaf3d-lychee. (a) The occlusion is not serious or there is no occlusion, (b) The situation of general occlusion, (c) and (d) are the detection results under severe occlusion.} 
\label{fig:graph8}
\end{figure*}

To determine the detection effect of the model in real environments, as well as the accuracy and robustness of the detection for different difficulties, this experiment was conducted in the case of no occlusion, average occlusion, and severe occlusion, respectively. For example, Fig. \ref{fig:graph8}(a) shows the case of no occlusion, Fig. \ref{fig:graph8}(b) shows the case of general occlusion, and Fig. \ref{fig:graph8}(c) and (d) shows the case of this gear in the eye. The detection performance of the three cases is shown in Table \ref{tab: Performance}, it can be seen that the accuracy of the picking point in the case of no occlusion is the highest, up to 0.932, and the highest accuracy of the position, in Fig. \ref{fig:graph8}(c) and (d) this more serious case of occlusion, the picking point of the gesture will have a greater error, and there will be interference to make the picking point of the false detection of the situation occurs, but the position of the point of the picking point is still very accurate, with an accuracy rate of 0.765. The reason is that under the occluded eye, the point cloud splicing is not very high quality for the picking point, which leads to the point cloud distortion at the picking point or the perception information is less, which makes it difficult for the model to mistakenly think that it is a picking point, resulting in a low accuracy rate.

\subsection{Discussion}
In this paper, a point cloud detection network based on deep learning is proposed to detect the pose of the target point in color point cloud data obtained by the depth camera sensor. The point cloud data of the same target at different three angles (frontal, two horizontal surfaces at 45-degree angles to the frontal) are stitched using the stitching algorithm, and a large number of points (200k-300k) after stitched are pre-processed by voxel down-sampling for efficient target detection of targets with different occlusion situations. The network detects the pose of the target point on the pre-processed point cloud data and guides the picking action of the robot arm.
\begin{figure}[ht]
\centering
\includegraphics[width=8.5cm]{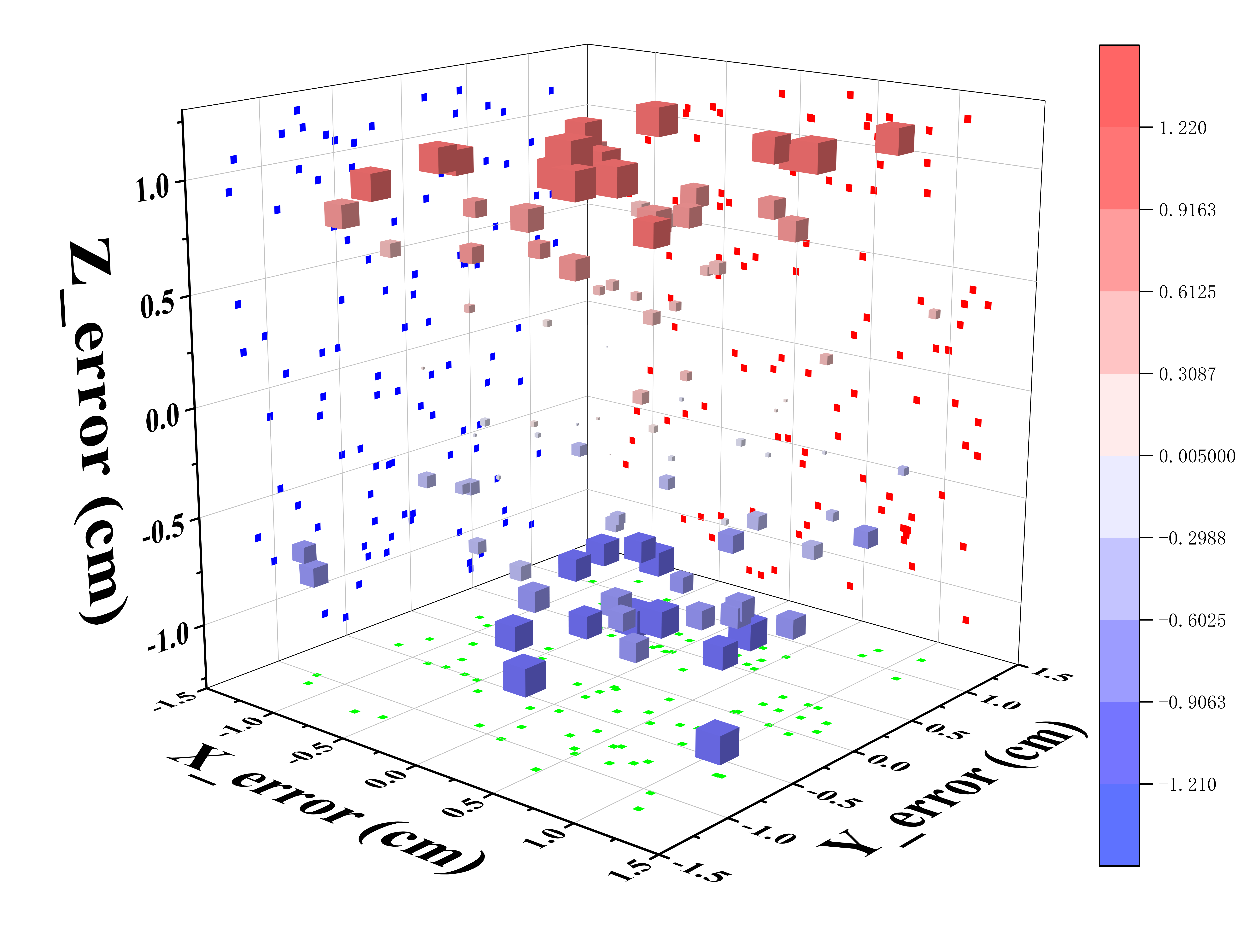}
\caption{3D localization errors of the lychee picking point. Each data point represented the error of that localization in three directions. The larger cube volume represented a larger localization error.} 
\label{fig:graph9}
\end{figure}

In the experiments, we analysed the performance impact of different downsampling sizes, attention mechanisms and the number of input channels of the residual block on the performance of the proposed Fcaf3d-lychee model. Experimental results show that the downsampling voxel size has a large impact on network performance. A voxel size that is too small (less than 0.01m) has no effect on improving network performance and is not beneficial for model resource usage. If the voxel size is too large (greater than 0.01m), the network performance may even be severely degraded due to unclear features. The experiment also verified the effect of the SE attention mechanism and the number of input channels of the residual block on the performance of the proposed model. Overall, with an appropriate choice of sampling voxel size, multi-viewpoint cloud stitching, attention mechanism fusion and number of input channels, the proposed Fcaf3d-lychee model achieves a $F_{1}$ score of 88.57\% on the test dataset. The target detection accuracy under different occlusion conditions in the real orchard environment reached 0.932, 0.824 and 0.765, respectively. To more simply express the 3D localisation error of Fcaf3d-lychee on the lychee picking point. We performed 100 localisation accuracy experiments in different scenarios. Then, the 3D error graph was plotted (Fig. \ref{fig:graph9}), which shows that the localisation errors in three directions are within the range of ±1.5 cm, reflecting the high localisation accuracy of the fcaf3d-lychee proposed in this paper.

\section{Conclusion} \label{section: conclusion}
This paper presents a novel method based on an enhanced 3D point cloud object detection network to accurately identify lychee-picking points in unstructured natural environments. The aim is to determine optimal picking points to guide picking robots to harvest lychee at suitable angles. Based on the Fcaf3d model, the research focuses on training lychee-picking point detection models to achieve high accuracy, robustness and generalisation.
The presented new detection network Fcaf3d-lychee, which incorporates attention and dynamic sparse pruning, shows excellent performance in various scenarios with an average accuracy (AP) of 83.39\% and a $F_{1}$ value of 88.57. Compared to other 3D detection models such as Tr3d and Votenet, the proposed network achieves significantly higher accuracy. Unlike traditional 2D detection networks such as YOLO, the network proposed in this paper directly obtains the 3D coordinates of lychee picking points, achieving effects that were previously unattainable.
Experimental results in real orchards show that the Fcaf3d-lychee model achieves robust detection performance under different shading conditions, with accuracies of 0.932, 0.824, and 0.765 for no shading, general shading, and heavy shading, respectively. Furthermore, the localisation accuracy of the Fcaf3d-lychee remains within ±1.5 cm in all directions, which is within the acceptable range of picking error.
The proposed Fcaf3d-lychee target detection network overcomes the limitations of previous methods, lays the foundation for the accurate calculation of picking points, and provides essential technical support for the practical application of lychee picking robots. 

\bibliographystyle{IEEEtran}
\bibliography{root}
\end{document}